\documentclass{article}

\usepackage{PRIMEarxiv}

\usepackage[utf8]{inputenc} 
\usepackage[T1]{fontenc}    
\usepackage{hyperref}       
\usepackage{url}            
\usepackage{booktabs}       
\usepackage{amsfonts}       
\usepackage{nicefrac}       
\usepackage{microtype}      
\usepackage{lipsum}
\usepackage{fancyhdr}       
\usepackage{graphicx}       
\graphicspath{{media/}}     

\title{DeNISE: Deep Networks for Improved Segmentation Edges}

\author{ \href{https://orcid.org/0009-0009-2798-9991}{\includegraphics[scale=0.06]{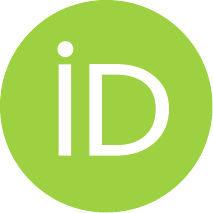}\hspace{1mm}Sander Riisøen Jyhne}\\
	Department of ICT\\
	University of Agder\\
	Grimstad, Norway \\
	\texttt{sander.jyhne@uia.no} \\
	\And
	\href{https://orcid.org/0000-0002-7742-4907}{\includegraphics[scale=0.06]{orcid.pdf}\hspace{1mm}Per-Arne Andersen} \\
	Department of ICT\\
	University of Agder\\
	Grimstad, Norway \\
	\texttt{per.andersen@uia.no} \\
        \And
	\href{https://orcid.org/0000-0001-6331-702X}{\includegraphics[scale=0.06]{orcid.pdf}\hspace{1mm}Morten Goodwin} \\
	Department of ICT\\
	University of Agder\\
	Grimstad, Norway \\
	\texttt{morten.goodwin@uia.no} \\
}

\begin{document}
\maketitle

\begin{abstract}
This paper presents Deep Networks for Improved Segmentation Edges (DeNISE), a novel data enhancement technique using edge detection and segmentation models to improve the boundary quality of segmentation masks. DeNISE utilizes the inherent differences in two sequential deep neural architectures to improve the accuracy of the predicted segmentation edge. DeNISE applies to all types of neural networks and is not trained end-to-end, allowing rapid experiments to discover which models complement each other. We test and apply DeNISE for building segmentation in aerial images. Aerial images are known for difficult conditions as they have a low resolution with optical noise, such as reflections, shadows, and visual obstructions. Overall the paper demonstrates the potential for DeNISE. Using the technique, we improve the baseline results with a building IoU of 78.9\%.
\end{abstract}


\section{Introduction}

Building segmentation is a prominent area of research in computer vision. A common problem for building segmentation is the quality and edge sharpness of the segmentation masks. Therefore, using predicted segmentation masks is limited to applications that do not rely on precise delineations of buildings. The precision of the current methods allows the detection of new or demolished buildings, thereby being valuable for change detection and disaster impact assessment applications. However, improved precision is necessary for applications with strict data quality requirements, such as map creation.

Public authorities, like the Norwegian Mapping Authority (NMA) \cite{kartverket}, use building data for map creation, policy-making, and city management \cite{Grecea2013CadastralPlanning}. NMA maintains a national building object database with essential information about each building. Furthermore, they update the database by manually measuring and annotating buildings in aerial imagery. Hence, using deep learning to annotate buildings reduces the cost and effort required to maintain an up-to-date database. With precise building masks, it is possible to incorporate building annotations into the building objects with significantly less human intervention than currently possible. Additionally, precise mask boundaries provide the basis for determining the building footprint, detecting ridgelines, and even creating 3D models of buildings.

Producing accurate segmentation masks is challenging because the training data derives from real-world data with varying quality. Optical factors, such as shadows, reflections, and perspectives, are present in aerial imagery and negatively influence the model's predictions. In addition, visibility is a challenge, as trees, powerlines, and other structures can block the view \cite{Schlosser2020BuildingSegmentation}. Despite many advancements in semantic segmentation, none of the models produce a segmentation boundary that satisfies map production standards.

However, many applications can successfully leverage building segmentation, including urban planning, disaster damage estimation, and change detection. The main reason is that the aforementioned applications do not require the same precision as map production. As a result, there is a research gap in the accuracy of the segmentation masks. Additionally, the models must generalize to various building types, shapes, and surrounding nature, ranging from urban cities to remote settlements.

To improve the precision of the predicted edges, we propose DeNISE, a technique highlighting buildings using an edge detection or segmentation network known as Edge-DeNISE and Seg-DeNISE, respectively. DeNISE combines the output prediction probabilities with the original images, directly modifying them or adding the predictions as a fourth image channel. A second network receives the modified images as input for training and prediction. 

The rest of the paper is structured as follows. Section \ref{sec:related_work} discusses related work to the proposed technique. Section \ref{sec:det} introduces the DeNISE method. Section \ref{sec:results}, presents the results and ablation studies. Finally, Section \ref{sec:conclusion} concludes the work and proposes paths for future work in enhanced segmentation boundaries. 

\section{Related Work}
\label{sec:related_work}
\subsection{Semantic Segmentation}
Semantic segmentation is finding and assigning the correct class to each pixel in an image, resulting in annotations of each class. Deep learning-based image segmentation research advances rapidly, with several new and novel architectures each year. One such architecture is the encoder-decoder architecture, which is proven successful for image segmentation tasks. U-Net \cite{Ronneberger2015U-net:Segmentation} is a well-known segmentation model using the encoder-decoder architecture with skip connections to retain the resolution through encoding and decoding. However, the skip connections are not able to retain all details. The HRNet \cite{Wang2021DeepRecognition} is an architecture mitigating this issue, a slightly different approach where three parallel and interconnected convolution streams maintain the high-resolution representations. However, the retained resolution comes at the cost of being more computationally expensive. Another popular approach for segmentation is dilated convolutions, which are used in other state-of-the-art segmentation models. Dilated convolutions are convolutions with holes between the convolution points, which allows for a larger convolution filter with reduced computational cost. DeeplabV3+ \cite{Chen2018Encoder-DecoderSegmentation} acquires state-of-the-art results on several tasks using this technique.

In recent years, the Transformer architecture has become popular for computer vision tasks. The authors of \cite{Dosovitskiy2021AnScale} propose the vision transformer leveraging attention mechanisms instead of convolutions. However, the attention mechanism of the transformer models is computationally expensive, limiting the resolution of the input image. The authors of \cite{Liu2022SwinResolution} introduce a shifted windows (SwinT) approach, reducing computation. The shifted windows lower the scale of self-attention by limiting it to non-overlapping local windows while still allowing cross-window attention, enabling global pixel context. Furthermore, in \cite{Li2022MaskSegmentation}, the authors propose a new transformer model building on SwinT to perform detection and segmentation in the same model, achieving state-of-the-art results.

\subsection{Edge detection}
Edge detection is the task of precisely delineating objects in an image. Deep-learning-based techniques are currently dominating the field, with several different methods available. One of the most famous is HED \cite{Xie2015Holistically-NestedDetection}, which uses a trimmed VGG16 backbone to generate multi-level features. Each layer produces a side output that is evaluated and backpropagated during training. A recent technique is the Bi-Directional Cascade Network (BDCN) \cite{He2019Bi-DirectionalDetection}. The authors argue that using the same supervision for all network layers is not optimal due to the different scales in each layer. They solve it by focusing shallow layers on details, while deep layers focus on object-level boundaries. However, BDCN struggles to detect crisp edge maps free of localization ambiguity due to the mixing phenomenon of CNNs. In 2021, \cite{Huan2021UnmixingDetection} introduces two novel modules; a tracing loss for feature unmixing and a fusion block for side mixing and aggregation of side edges. The modules can be paired with other edge detection models, such as BDCN, increasing the performance. 

\subsection{Aerial Image Segmentation}
A common approach for aerial image segmentation is applying state-of-the-art segmentation models and training them to segment buildings, roads, or other objects. Among them, \cite{Pan2020DeepU-Net}, \cite{Zhang2018RoadU-Net}, and \cite{Zhao2019UseLodging} have all used a U-Net architecture with relatively good results. Other approaches modify existing architectures to extract specific object features. \cite{Ghandorh2022SemanticImages} propose a technique to improve the segmentation masks of roads by utilizing edge detection. The model segments the images and passes them to an edge detection network with the encoded image. As a result, the edge detection module relies on the performance of the segmentation network. Additionally, the model is complex, and changing the edge detection part is not easy. Similarly, \cite{Lee2022Boundary-OrientedImages} also approaches the problem using two stages in the model. However, they utilize two segmentation networks in a two-scheme method using a novel boundary loss and a USIM module. The predictions of the first network are combined with the original image in the USIM module and passed through the second network. The method improves upon using one segmentation network. The complexity of the model makes it difficult to swap the existing models and causes the training to be resource-intensive. Lastly, they display the improvements of their method using older models that are not state-of-the-art.

\section{Deep Networks for Improved Segmentation Edges (DeNISE)}
\label{sec:det}
\begin{figure*}[ht]
    \centering\includegraphics[width=1\linewidth]{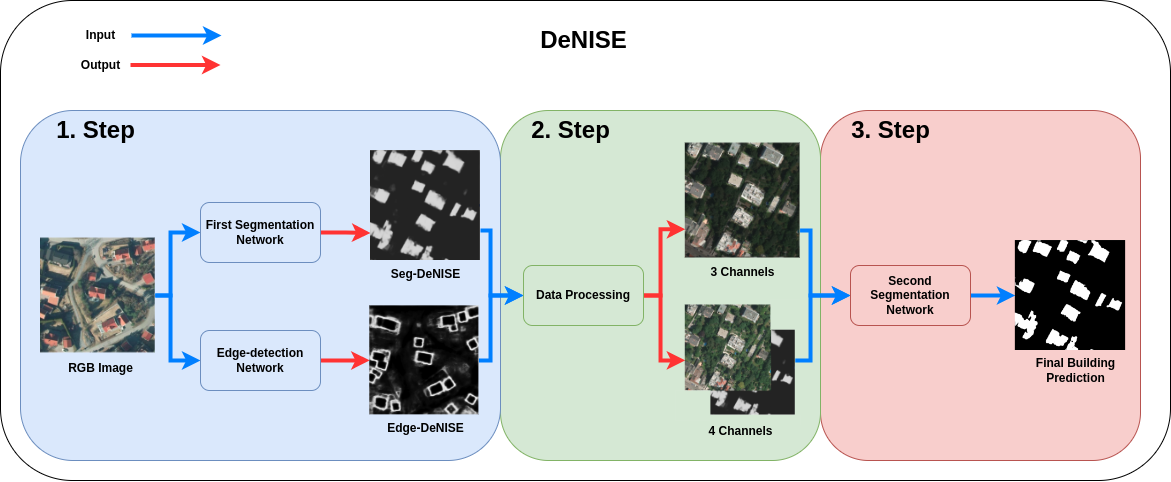}
    \caption{A visual representation of DeNISE illustrating the process of combining the predictions of a neural network with the original image data in three steps. (1) A segmentation or edge detection model trains on the unenhanced dataset. (2) DeNISE generates training data by enhancing the original data with the predictions from the first network. (3) The second network uses the enhanced dataset for training and inference. The branches in the figure represent a choice between available methods at the current step.}
    \label{fig:generaldet}
\end{figure*}

The primary objective of DeNISE is to improve the boundary quality of segmentation masks for buildings compared to standalone segmentation models. Figure \ref{fig:generaldet} depicts a high-level overview of DeNISE.

Both DeNISE approaches use the same structure, combining the strengths of two different neural networks to improve the precision of predicted edges. Because the architecture decouples both models, the user can easily swap the first and second neural networks with other models. The ability to quickly try different models allows rapid testing to find complementary models. 

We propose two DeNISE approaches: (1) Seg-DeNISE, which uses two segmentation networks, and (2) Edge-DeNISE, using an edge detection network in conjunction with a segmentation network. In both approaches, the first model performs inference on the original training data, then combines the predictions with the data and uses the combination for training and inference in the second model.

\subsection{Seg-DeNISE}
\begin{figure*}[ht]
    \centering\includegraphics[width=0.5\linewidth]{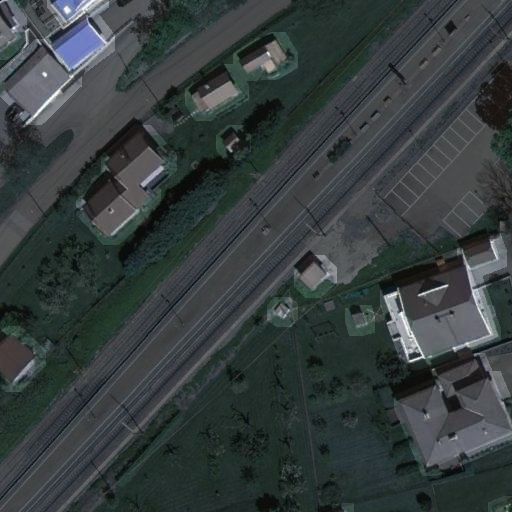}
    \caption{An example of the enhanced data from 3-channel Seg-DeNISE, showing the dilated region surrounding the buildings.}
    \label{fig:3channel}
\end{figure*}

The first segmentation network in Seg-DeNISE receives an input image and outputs the class probabilities. Seg-DeNISE can use the probabilities in two ways: they can be concatenated with the original input or merged into the image while retaining the original three channels. The merging consists of four steps to create the enhanced data. (1) Threshold the probabilities from the first neural network. We set all gradients equal to or greater than 0.5 to 1 and the rest to 0. (2) The thresholded predictions are dilated 15 pixels in each direction to ensure the mask covers the entire building. (3) All predictions are clipped values between 0.5 and 1.0. This step is crucial to include the background information when multiplying the prediction gradients with the original image. Additionally, clipping the gradients allow the second segmentation to find buildings that the first model missed. (4) The dilated and clipped predictions are multiplied by the original input image. The result shown in Figure \ref{fig:3channel} is an image where all dilated building masks keep the original brightness while the rest is set to 50\% brightness.

\subsection{Edge-DeNISE}
The edge detection network in Edge-DeNISE receives an input image and predicts the corresponding building edge gradients. We use the predicted gradients in two ways. The gradients can be concatenated with the original input or merged into the image while retaining the three channels. The merging process for Edge-DeNISE is similar to Seg-DeNISE, using two steps. (1) The predicted gradients are thresholded and clipped between 0.5 and 1.0, preserving the background information. (2) The clipped gradients are multiplied with the original input image, keeping the original brightness on the predicted building edges while setting the rest of the image to 50\% brightness.


\subsection{NMA Dataset}
We use a private dataset named NMA dataset for the experiments. The dataset consists of orthophotos and ground truth masks for buildings in Norway. The orthophotos are created using a Digital Terrain Model (DTM), which depicts the height of the terrain in a given area. All points in orthophotos created using a DTM are slightly skewed. As a result, the ground truth masks and buildings depicted in the orthophotos do not perfectly align. The dataset is divided into training, validation, and test splits with 22 892, 2861, and 2862 images, respectively.

\section{Results}
\label{sec:results}
We evaluate Seg-DeNISE and Egde-DeNISE on the NMA dataset. For all experiments, we use three different segmentation models, and an edge detection model, namely Hierarchical Multi-Scale Attention (HMSA) \cite{Tao2020HierarchicalSegmentation}, DeeplabV3+ (DL3) \cite{Chen2018Encoder-DecoderSegmentation}, U-Net \cite{Ronneberger2015U-net:Segmentation}, and CATS-BDCN (CATS) \cite{Huan2021UnmixingDetection}, respectively. For all experiments, the models train for 20 epochs with a batch size of 8 and a learning rate of 1e-4. 

Furthermore, Seg-DeNISE always uses DL3 as the first segmentation network in Seg-DeNISE and evaluates using all models as the second segmentation network. Edge-DeNISE uses CATS as the first network and, similarly to Seg-DeNISE, evaluates using all segmentation models as the second network. None of the experiments utilize pre-trained encoders, mainly due to the lack of pre-trained encoders for 4-channel data, making a fair comparison difficult. For the evaluation of experiments, we use Intersection-over-Union (IoU) and Boundary IoU (BIoU) \cite{Cheng2021BoundaryEvaluation}, an IoU measure focusing on the predicted segmentation edges.

\subsection{Baseline}
 We run experiments with all segmentation models as a standalone network, creating a baseline for comparison. Table \ref{tab:results} display the baseline evaluation results, establishing the superiority of the HMSA model, followed by DL3 and U-Net.

\begin{table}
  \footnotesize
  \begin{center}
    \caption{Evaluation results on the NMA dataset for all experiments. The results show DeNISE increasing the IoU and BIoU measures for the U-Net and DL3 models. However, it does not outperform the HMSA baseline.}
    \label{tab:results}
    \begin{tabular}{ccccccc}
      \toprule
      \emph{Model} & \emph{Dataset} & \emph{Loss} & \emph{Method} & \emph{IoU} & \emph{BIoU} & \emph{Source}\\
      \midrule
      U-Net (Baseline) & NMA & RMI & Standalone & 0.7657 & 0.6279 & \cite{Ronneberger2015U-net:Segmentation}\\
      \hline
      U-Net & NMA & RMI & Seg-DeNISE (3-channels) & 0.7685 & 0.6343 & Ours\\
      \hline
      U-Net & NMA & RMI & Seg-DeNISE (4-channels) & 0.7730 & 0.6416 & Ours\\
      \hline
      \textbf{U-Net} & \textbf{NMA} & \textbf{RMI} & \textbf{Edge-DeNISE (3-channels)} & \textbf{0.7742} & \textbf{0.6445} & \textbf{Ours}\\
      \hline
      U-Net & NMA & RMI & Edge-DeNISE (4-channels) & 0.7739 & 0.6480 & Ours\\
      \toprule
      DL3 (Baseline) & NMA & RMI & Standalone & 0.7850 & 0.6586 & \cite{Chen2018Encoder-DecoderSegmentation}\\
      \hline
      DL3 & NMA & RMI & Seg-DeNISE (3-channels) & 0.7802 & 0.6535 & Ours\\
      \hline
      DL3 & NMA & RMI & Seg-DeNISE (4-channels) & 0.7778 & 0.6492 & Ours\\
      \hline
      DL3 & NMA & RMI & Edge-DeNISE (3-channels) & 0.7882 & 0.6658 & Ours\\
      \hline
      \textbf{DL3} & \textbf{NMA} & \textbf{RMI} & \textbf{Edge-DeNISE (4-channels)} & \textbf{0.7890} & \textbf{0.6691} & \textbf{Ours}\\
      \toprule
      \textbf{HMSA (Baseline)} & \textbf{NMA} & \textbf{RMI} & \textbf{Standalone} & \textbf{0.8291} & \textbf{0.7284} & \textbf{\cite{Tao2020HierarchicalSegmentation}}\\
      \hline
      HMSA & NMA & RMI & Seg-DeNISE (3-channels) & 0.8103 & 0.7003 & Ours\\
      \hline
      HMSA & NMA & RMI & Seg-DeNISE (4-channels) & 0.7997 & 0.6860 & Ours\\
      \hline
      HMSA & NMA & RMI & Edge-DeNISE (3-channels) & 0.8076 & 0.6968 & Ours\\
      \hline
      HMSA & NMA & RMI & Edge-DeNISE (4-channels) & 0.8057 & 0.6975 & Ours\\
      \bottomrule
    \end{tabular}
  \end{center}
\end{table}

\subsection{Seg-DeNISE}
Table \ref{tab:results} presents the results for Seg-DeNISE, revealing subpar evaluation scores where U-Net is the only model that improves upon the baseline. The evaluation results suggest that a relatively large gap in performance between the first and second segmentation models may negatively influence the accuracy of the predictions. Additionally, the evaluation scores indicate that the second segmentation network cannot correct the mistakes of the first network. Lastly, comparing the results for 3 and 4 channels, a slight trend in performance is present, where 3 channels have the best evaluation scores by a small margin.

\subsection{Edge-DeNISE}
Table \ref{tab:results} reveals promising evaluation scores for Edge-DeNISE, as U-Net and DL3 improve upon the baselines. The results indicate that the edge detection network's performance relative to the segmentation network's is crucial for improving the results. Furthermore, the results suggest that the predictions from CATS for HMSA introduce noise instead of enhancing the data, negatively impacting the results. Dissimilar to Seg-DeNISE, the Boundary IoU scores for Edge-DeNISE slightly favor 4 channels compared to 3.

\section{Conclusion}
\label{sec:conclusion}
This paper proposes DeNISE, a data enhancement technique that uses either a segmentation (Seg-DeNISE) or an edge detection (Edge-DeNISE) network to enhance the original data using their predictions. Subsequently, a secondary segmentation network uses the enhanced data for training and inference, improving the precision of predicted segmentation masks. Our results for Seg-DeNISE show inconsistent performance, indicating that the first segmentation network can negatively or positively impact Seg-DeNISE results. However, Edge-DeNISE yields promising results and improves upon the baseline results for two out of three models. The results suggest that further advancement of edge detection literature will greatly benefit Edge-DeNISE.


\subsection{Future Work}
\label{sec:future_work}
This paper proposes a new approach for improving the edges of the predicted segmentation masks. The results advocate further investigation into using neural networks to enhance the training data for a second neural network. Several new but potential directions for DeNISE need further research. 

\begin{itemize}
    \item Using an object detection model as the first network.
    \item Explore other combinations of segmentation and edge detection models for DeNISE.
    \item Improving the edge detection literature to further improve Edge-DeNISE.
    \item Training two segmentation models end-to-end, where the latent space from the first model is merged with the latent space of the second model.
    \item Reversing Edge-DeNISE by having the segmentation network first and edge detection last.
\end{itemize}


\begin{thebibliography}{10}

\bibitem{kartverket}
{The Norwegian Mapping Authority}.
\newblock {The Norwegian Mapping Authority}, 9 2022.

\bibitem{Grecea2013CadastralPlanning}
Carmen Grecea, Alina B{\u{a}}l{\u{a}}, and S~Herban.
\newblock {Cadastral requirements for urban administration, key component for
  an efficient town planning}.
\newblock {\em Journal of Environmental Protection and Ecology}, 14:363--371,
  10 2013.

\bibitem{Schlosser2020BuildingSegmentation}
Aletta~Dóra Schlosser, Gergely Szab{\'{o}}, László Bertalan, Zsolt Varga,
  Péter Enyedi, and Szilárd Szab{\'{o}}.
\newblock {Building Extraction Using Orthophotos and Dense Point Cloud Derived
  from Visual Band Aerial Imagery Based on Machine Learning and Segmentation}.
\newblock {\em Remote Sensing 2020, Vol. 12, Page 2397}, 12(15):2397, 7 2020.

\bibitem{Ronneberger2015U-net:Segmentation}
Olaf Ronneberger, Philipp Fischer, and Thomas Brox.
\newblock {U-net: Convolutional networks for biomedical image segmentation}.
\newblock {\em Lecture Notes in Computer Science (including subseries Lecture
  Notes in Artificial Intelligence and Lecture Notes in Bioinformatics)},
  9351:234--241, 2015.

\bibitem{Wang2021DeepRecognition}
Jingdong Wang, Ke~Sun, Tianheng Cheng, Borui Jiang, Chaorui Deng, Yang Zhao,
  Dong Liu, Yadong Mu, Mingkui Tan, Xinggang Wang, Wenyu Liu, and Bin Xiao.
\newblock {Deep High-Resolution Representation Learning for Visual
  Recognition}.
\newblock {\em IEEE Transactions on Pattern Analysis and Machine Intelligence},
  43(10):3349--3364, 10 2021.

\bibitem{Chen2018Encoder-DecoderSegmentation}
Liang~Chieh Chen, Yukun Zhu, George Papandreou, Florian Schroff, and Hartwig
  Adam.
\newblock {Encoder-Decoder with Atrous Separable Convolution for Semantic Image
  Segmentation}.
\newblock {\em Lecture Notes in Computer Science (including subseries Lecture
  Notes in Artificial Intelligence and Lecture Notes in Bioinformatics)}, 11211
  LNCS:833--851, 2 2018.

\bibitem{Dosovitskiy2021AnScale}
Alexey Dosovitskiy, Lucas Beyer, Alexander Kolesnikov, Dirk Weissenborn,
  Xiaohua Zhai, Thomas Unterthiner, Mostafa Dehghani, Matthias Minderer, Georg
  Heigold, Sylvain Gelly, Jakob Uszkoreit, and Neil Houlsby.
\newblock {An Image is Worth 16x16 Words: Transformers for Image Recognition at
  Scale}.
\newblock In {\em Proceedings of the 9th International Conference on Learning
  Representations (ICLR)}, pages 1--21, 10 2021.

\bibitem{Liu2022SwinResolution}
Ze~Liu, Han Hu, Yutong Lin, Zhuliang Yao, Zhenda Xie, Yixuan Wei, Jia Ning, Yue
  Cao, Zheng Zhang, Li~Dong, Furu Wei, and Baining Guo.
\newblock {Swin Transformer V2: Scaling Up Capacity and Resolution}, 2022.

\bibitem{Li2022MaskSegmentation}
Feng Li, Hao Zhang, Huaizhe xu, Shilong Liu, Lei Zhang, Lionel~M. Ni, and
  Heung-Yeung Shum.
\newblock {Mask DINO: Towards A Unified Transformer-based Framework for Object
  Detection and Segmentation}.
\newblock 6 2022.

\bibitem{Xie2015Holistically-NestedDetection}
Saining Xie and Zhuowen Tu.
\newblock {Holistically-Nested Edge Detection}, 2015.

\bibitem{He2019Bi-DirectionalDetection}
Jianzhong He, Shiliang Zhang, Ming Yang, Yanhu Shan, and Tiejun Huang.
\newblock {Bi-Directional Cascade Network for Perceptual Edge Detection}, 2019.

\bibitem{Huan2021UnmixingDetection}
Linxi Huan, Nan Xue, Xianwei Zheng, Wei He, Jianya Gong, and Gui~Song Xia.
\newblock {Unmixing Convolutional Features for Crisp Edge Detection}.
\newblock {\em IEEE Transactions on Pattern Analysis and Machine Intelligence},
  2021.

\bibitem{Pan2020DeepU-Net}
Zhuokun Pan, Jiashu Xu, Yubin Guo, Yueming Hu, and Guangxing Wang.
\newblock {Deep Learning Segmentation and Classification for Urban Village
  Using a Worldview Satellite Image Based on U-Net}.
\newblock {\em Remote Sensing 2020, Vol. 12, Page 1574}, 12(10):1574, 5 2020.

\bibitem{Zhang2018RoadU-Net}
Zhengxin Zhang, Qingjie Liu, and Yunhong Wang.
\newblock {Road Extraction by Deep Residual U-Net}.
\newblock {\em IEEE Geoscience and Remote Sensing Letters}, 15(5):749--753, 5
  2018.

\bibitem{Zhao2019UseLodging}
Xin Zhao, Yitong Yuan, Mengdie Song, Yang Ding, Fenfang Lin, Dong Liang, and
  Dongyan Zhang.
\newblock {Use of Unmanned Aerial Vehicle Imagery and Deep Learning UNet to
  Extract Rice Lodging}.
\newblock {\em Sensors 2019, Vol. 19, Page 3859}, 19(18):3859, 9 2019.

\bibitem{Ghandorh2022SemanticImages}
Hamza Ghandorh, Wadii Boulila, Sharjeel Masood, Anis Koubaa, Fawad Ahmed, and
  Jawad Ahmad.
\newblock {Semantic Segmentation and Edge Detection{\&}mdash;Approach to Road
  Detection in Very High Resolution Satellite Images}.
\newblock {\em Remote Sensing 2022, Vol. 14, Page 613}, 14(3):613, 1 2022.

\bibitem{Lee2022Boundary-OrientedImages}
Kyungsu Lee, Jun~Hee Kim, Haeyun Lee, Juhum Park, Jihwan~P. Choi, and Jae~Youn
  Hwang.
\newblock {Boundary-Oriented Binary Building Segmentation Model with Two Scheme
  Learning for Aerial Images}.
\newblock {\em IEEE Transactions on Geoscience and Remote Sensing}, 60, 2022.

\bibitem{Tao2020HierarchicalSegmentation}
Andrew Tao, Karan Sapra, and Bryan Catanzaro.
\newblock {Hierarchical Multi-Scale Attention for Semantic Segmentation}.
\newblock {\em CoRR}, abs/2005.10821, 2020.

\bibitem{Cheng2021BoundaryEvaluation}
Bowen Cheng, Ross Girshick, Piotr Dollar, Alexander~C. Berg, and Alexander
  Kirillov.
\newblock {Boundary IoU: Improving Object-Centric Image Segmentation
  Evaluation}, 2021.

\end{thebibliography}

\end{document}